\RequirePackage{fix-cm}
\documentclass[smallextended]{svjour3}       
\smartqed  
\usepackage{graphicx}
\usepackage{algpseudocode}
\usepackage[ruled, lined, linesnumbered, commentsnumbered, longend]{algorithm2e}

\begin{document}

\title{Attention Mechanism in Neural Networks:
}
\subtitle{Where it Comes and Where it Goes}


\author{Derya Soydaner
}


\institute{Derya Soydaner \at
              Department of Brain and Cognition, University of Leuven (KU Leuven), Leuven, Belgium \\
              Tel.: +32-16710471\\
              \email{derya.soydaner@kuleuven.be}           
}

\date{Received: 22 July 2021 / Accepted: 27 April 2022}

\maketitle

\begin{abstract}
A long time ago in the machine learning literature, the idea of incorporating a mechanism inspired by the human visual system into neural networks was introduced. This idea is named the \emph{attention mechanism}, and it has gone through a long development period. Today, many works have been devoted to this idea in a variety of tasks. Remarkable performance has recently been demonstrated. The goal of this paper is to provide an overview from the early work on searching for ways to implement attention idea with neural networks until the recent trends. This review emphasizes the important milestones during this progress regarding different tasks. By this way, this study aims to provide a road map for researchers to explore the current development and get inspired for novel approaches beyond the attention.  
\end{abstract}

\section{Introduction}
\label{intro}
Human eye sees the world in an interesting way. We suppose as if we see the entire scene at once, but this is an illusion created by the subconscious part of our brain \cite{goodfellow2016}. According to the {\em Scanpath} theory \cite{noton1971,noton1971a}, when the human eye looks at an image, it can see only a small patch in high resolution. This small patch is called the \emph{fovea}. It can see the rest of the image in low resolution which is called the \emph{periphery}. To recognize the entire scene, the eye performs feature extraction based on the fovea. The eye is moved to different parts of the image until the information obtained from the fovea is sufficient for recognition \cite{alpaydin1996}. These eye movements are called \emph{saccades}. The eye makes successive fixations until the recognition task is complete. This sequential process happens so quickly that we feel as if it happens all at once. 

Biologically, this is called \emph{visual attention system}. Visual attention is defined as the ability to dynamically restrict processing to a subset of the visual field \cite{ahmad1991}. It seeks answers for two main questions: \emph{What} and \emph{where} to look? Visual attention has been extensively studied in psychology and neuroscience; for reviews see \cite{posner1990,bundesen1990,desimone1995,corbetta2002,petersen2012}. Besides, there is a large amount of literature on modeling eye movements \cite{rimey1990,sheliga1994,sheliga1995,hoffman1995}. These studies have been a source of inspiration for many artificial intelligence tasks. It has been discovered that the attention idea is useful from image recognition to machine translation. Therefore, different types of attention mechanisms inspired from the human visual system have been developed for years. Since the success of deep neural networks has been at the forefront for these artificial intelligence tasks, these mechanisms have been integrated into neural networks for a long time.  

This survey is about the journey of attention mechanisms used with neural networks. Researchers have been investigating ways to strengthen neural network architectures with attention mechanisms for many years. The primary aim of these studies is to reduce computational burden and to improve the model performance as well. Previous work reviewed the attention mechanisms from different perspectives \cite{chaudhari2019}, or examined them in context of natural language processing (NLP) \cite{galassi2019,lee2019a}. However, in this study, we examine the development of attention mechanisms over the years, and recent trends. We begin with the first attempts to integrate the visual attention idea to neural networks, and continue until the most modern neural networks armed with attention mechanisms. One of them is the \emph{Transformer}, which is used for many studies including the \emph{GPT-3} language model \cite{brown2020}, goes  beyond convolutions and recurrence by replacing them with only attention layers \cite{vaswani2017}. Finally, we discuss how much more can we move forward, and what's next?

\section{From the Late 1980s to Early 2010s: The Attention Awakens}
\label{sec:1}
The first attempts at adapting attention mechanisms to neural networks go back to the late 1980s. One of the early studies is the improved version of the \emph{Neocognitron} \cite{fukushima1980} with selective attention \cite{fukushima1987}. This study is then modified to recognize and segment connected characters in cursive handwriting \cite{fukushima1993}. Another study describes {\em VISIT}, a novel model that concentrates on its relationship to a number of visual areas of the brain \cite{ahmad1991}. Also, a novel architecture named {\em Signal Channelling Attentional Network (SCAN)} is presented for attentional scanning \cite{postma1997}.  

Early work on improving the attention idea for neural networks includes a variety of tasks such as target detection \cite{schmidhuber1991}. In another study, a visual attention system extracts regions of interest by combining the bottom-up and top-down information from the image \cite{milanese1994}.  A recognition model based on selective attention which analyses only a small part of the image at each step, and combines results in time is described \cite{alpaydin1996}. Besides, a model based on the concept of selective tuning is proposed \cite{tsotsos1995}. As the years go by, several studies that use the attention idea in different ways have been presented for visual perception and recognition \cite{culhane1992,reisfeld1995,rybak1998,keller1999}. 

By the 2000s, the studies on making attention mechanisms more useful for neural networks continued. In the early years, a model that integrates an attentional orienting \emph{where} pathway and an object recognition \emph{what} pathway is presented \cite{miau2001}. A computational model of human eye movements is proposed for an object class detection task \cite{Zhang2006}. A serial model is presented for visual pattern recognition gathering Markov models and neural networks with selective attention on the handwritten digit recognition and face recognition problems \cite{salah2002}. In that study, a neural network analyses image parts and generates posterior probabilities as observations to the Markov model. Also, attention idea is used for object recognition \cite{walther2002}, and the analysis of a scene \cite{schill2001}. An interesting study proposes to learn sequential attention in real-world visual object recognition using a Q-learner \cite{paletta2005}. Besides, a computational model of visual selective attention is described to automatically detect the most relevant parts of a color picture displayed on a television screen \cite{lemeur2006}. The attention idea is also used for identifying and tracking objects in multi-resolution digital video of partially cluttered environments \cite{gould2007}.

In 2010, the first implemented system inspired by the fovea of human retina was presented for image classification \cite{larochelle2010}. This system jointly trains a restricted Boltzmann machine (RBM) and an attentional component called {\em the fixation controller}. Similarly, a novel attentional model is implemented for simultaneous object tracking and recognition that is driven by gaze data \cite{bazzani2011}. By taking advantage of reinforcement learning, a novel recurrent neural network (RNN) is described for image classification \cite{mnih2014}. \emph {Deep Attention Selective Network (DasNet)}, a deep neural network with feedback connections that are learned through reinforcement learning to direct selective attention to certain features extracted from images, is presented \cite{stollenga2014}. Additionally, a deep learning based framework using attention has been proposed for generative modeling \cite{tang2014}.
\section{2015: The Rise of Attention}
\label{sec:2}
It can be said that 2015 is the golden year of attention mechanisms. Because the number of attention studies has grown like an avalanche after three main studies presented in that year. The first one proposed a novel approach for neural machine translation (NMT) \cite{bahdanau2015}. As it is known, most of the NMT models belong to a family of encoder-decoders \cite{sutskever2014,cho2014a}, with an encoder and a decoder for each language. However, compressing all the necessary information of a source sentence into a fixed-length vector is an important disadvantage of this encoder-decoder approach. This usually makes it difficult for the neural network to capture all the semantic details of a very long sentence \cite{goodfellow2016}.

\begin{figure}[h!]
\centering
\includegraphics[scale=.5]{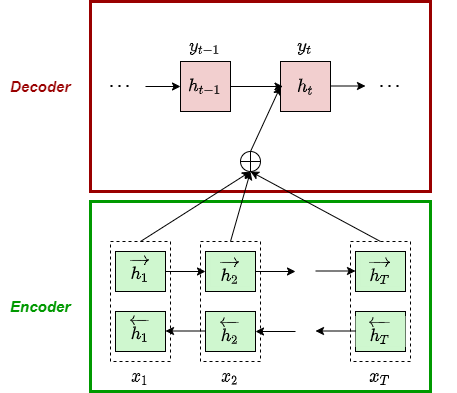}
\caption{The extension to the conventional NMT models that is proposed by \cite{bahdanau2015}. It generates the \emph{t}-th target word $y_t$ given a source sentence $(x_1, x_2, ..., x_T)$. }
\label{encoder_decoder}
\end{figure}

The idea that \cite{bahdanau2015} introduced is an extension to the conventional NMT models. This extension is composed of an encoder and decoder as shown in Fig \ref{encoder_decoder}. The first part, encoder, is a {\em bidirectional RNN} (BiRNN) \cite{schuster1997} that takes word vectors as input. The forward and backward states of BiRNN are computed. Then, an \emph{annotation} $a_j$ for each word $x_j$ is obtained by concatenating these forward and backward hidden states. Thus, the encoder maps the input sentence to a sequence of annotations $(a_1,...,a_{T_x})$. By using a BiRNN rather than conventional RNN, the annotation of each word can summarize both the preceding words and the following words. Besides, the annotation $a_j$ can focus on the words around $x_j$ because of the inherent nature of RNNs that representing recent inputs better.    

In decoder, a weight $\alpha_{ij}$ of each annotation $a_j$ is obtained by using its associated energy $e_{ij}$ that is computed by a feedforward neural network \emph{f} as in Eq. (\ref{bahdanauenergy}). This neural network \emph{f} is defined as an \emph{alignment model} that can be jointly trained with the proposed architecture. In order to reduce computational burden, a multilayer perceptron (MLP) with a single hidden layer is proposed as \emph{f}. This alignment model tells us about the relation between the inputs around position \emph{j} and the output at position \emph{i}. By this way, the decoder applies an attention mechanism. As it is seen in Eq. (\ref{bahdanaualpha}), the $\alpha_{ij}$ is the output of softmax function:

\begin{equation}
e_{ij} = f(h_{i-1},a_j)
\label{bahdanauenergy}
\end{equation}

\begin{equation}
\alpha_{ij} = \frac{\exp(e_{ij})}{\sum_{k=1}^{T_x}\exp(e_{ik})}
\label{bahdanaualpha}
\end{equation}

Here, the probability $\alpha_{ij}$ determines the importance of annotation $a_j$ with respect to the previous hidden state $h_{i-1}$. Finally, the context vector $c_i$ is computed as a weighted sum of these annotations as follows \cite{bahdanau2015}: 

\begin{equation}
c_i = \sum_{j=1}^{T_x} \alpha_{ij} a_j
\label{bahdanauattention}
\end{equation}

Based on the decoder state, the context and the last generated word, the target word $y_t$ is predicted. In order to generate a word in a translation, the model searches for the most relevant information in the source sentence to concentrate. When it finds the appropriate source positions, it makes the prediction. By this way, the input sentence is encoded into a sequence of vectors and a subset of these vectors is selected adaptively by the decoder that is relevant to predicting the target \cite{bahdanau2015}. Thus, it is no longer necessary to compress all the information of a source sentence into a fixed-length vector. 

The second study is the first visual attention model in image captioning \cite{xu2015}. Different from the previous study \cite{bahdanau2015}, it uses a deep convolutional neural network (CNN) as an encoder. This architecture is an extension of the neural network \cite{vinyals2015} that encodes an image into a compact representation, followed by an RNN that generates a corresponding sentence. Here, the {\em annotation vectors} $a_i \in R^D$ are extracted from a lower convolutional layer, each of which is a \emph{D}-dimensional representation corresponding to a part of the image. Thus, the decoder selectively focuses on certain parts of an image by weighting a subset of all the feature vectors \cite{xu2015}. This extended architecture uses attention for salient features to dynamically come to the forefront instead of compressing the entire image into a static representation. 

The context vector $c_t$ represents the relevant part of the input image at time \emph{t}. The weight $\alpha_i$ of each annotation vector is computed similar to Eq. (\ref{bahdanaualpha}), whereas its associated energy is computed similar to Eq. (\ref{bahdanauenergy}) by using an MLP conditioned on the previous hidden state $h_{t-1}$. The remarkable point of this study is a new mechanism  $\phi$ that computes $c_t$ from the annotation vectors $a_i$ corresponding to the features extracted at different image locations: 

\begin{equation}
c_t = \phi(\big\{a_i\big\},\big\{\alpha_i\big\})
\end{equation}

The definition of the $\phi$ function causes two variants of attention mechanisms: The \emph{hard (stochastic)} attention mechanism is trainable by maximizing an approximate variational lower bound, i.e., by REINFORCE \cite{williams1992}. On the other side, the \emph{soft (deterministic)} attention mechanism is trainable by standard backpropagation methods. The hard attention defines a location variable $s_t$, and uses it to decide where to focus attention when generating the \emph{t-th} word. When the hard attention is applied, the attention locations are considered as intermediate latent variables. It assigns a multinoulli distribution parametrized by ${\alpha_i}$, and $c_t$ becomes a random variable. Here, $s_{t,i}$ is defined as a one-hot variable which is set to 1 if the \emph{i-th} location is used to extract visual features \cite{xu2015}:

\begin{equation}
p(s_{t,i} = 1 | s_{j<t}, a) = \alpha_{t,i}
\end{equation}    

\begin{equation}
c_t = \sum_i s_{t,i} a_i 
\end{equation}

Whereas learning hard attention requires sampling the attention location $s_t$ each time, the soft attention mechanism computes a weighted annotation vector similar to \cite{bahdanau2015} and takes the expectation of the context vector $c_t$ directly:

\begin{equation}
E_{p(s_t|\alpha)}[c_t] = \sum_{i=1}^L \alpha_{t,i} a_i
\end{equation}

Furthermore, in training the deterministic version of the model, an alternative method namely \emph{doubly stochastic attention}, is proposed with an additional constraint added to the training objective to encourage the model to pay equal attention to all parts of the image.

The third study should be emphasized presents two classes of attention mechanisms for NMT: the \emph{global} attention that always attends to all source words, and the \emph{local} attention that only looks at a subset of source words at a time \cite{luong2015}. These mechanisms derive the context vector $c_t$ in different ways: Whereas the global attention considers all the hidden states of the encoder, the local one selectively focuses on a small window of context. In global attention, a variable-length alignment vector is derived similar to Eq. (\ref{bahdanaualpha}). Here, the current target hidden state $h_t$ is compared with each source hidden state $\bar{h}_s$ by using a \emph{score} function instead of the associated energy $e_{ij}$. Thus, the alignment vector whose size equals the number of time steps on the source side is derived.  
Given the alignment vector as weights, the context vector $c_t$ is computed as the weighted average over all the source hidden states. Here, \emph{score} is referred as a \emph{content-based} function, and three different alternatives are considered \cite{luong2015}.

On the other side, the local attention is differentiable. Firstly, an aligned position $p_t$ is generated for each target word at a time $t$. Then, a window centered around the source position $p_t$ is used to compute the context vector as a weighted average of the source hidden states within the window. The local attention selectively focuses on a small window of context, and obtains the alignment vector from the current target state $h_t$ and the source states $\bar{h}_s$ in the window \cite{luong2015}.  

\begin{figure}[h!]
\centering
\includegraphics[scale=.6]{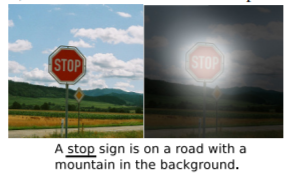}\\
\includegraphics[scale=.6]{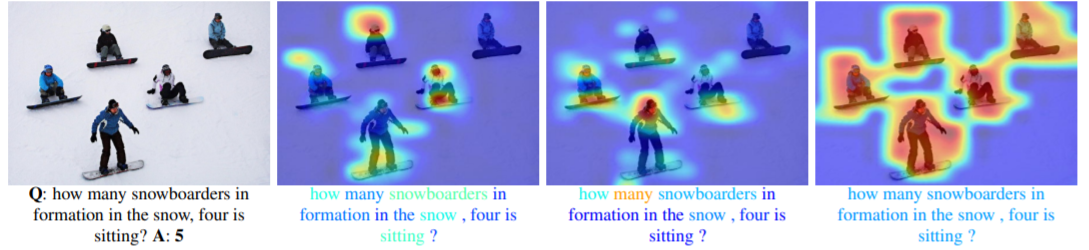}
\caption{Examples of the attention mechanism in visual. \emph{(Top)} Attending to the correct object in neural image caption generation \cite{xu2015}. \emph{(Bottom)} Visualization of original image and question pairs, and co-attention maps namely word-level, phrase-level and question-level, respectively \cite{lu2017a}.}
\label{local attentional model}
\end{figure}

The introduction of these novel mechanisms in 2015 triggered the rise of attention for neural networks. Based on the proposed attention mechanisms, significant research has been conducted in a variety of tasks. In order to imagine the attention idea in neural networks better, two visual examples are shown in Fig. \ref{local attentional model}. A neural image caption generation task is seen in the top row that implements an attention mechanism \cite{xu2015}. Then, the second example shows how the attention mechanisms can be used for visual question answering \cite{lu2017a}. Both examples demonstrate how attention mechanisms focus on parts of input images.

\section{2015-2016: Attack of the Attention}
\label{sec:3}
During two years from 2015, the attention mechanisms were used for different tasks, and novel neural network architectures were presented applying these mechanisms. After the \emph{memory networks} \cite{weston2014} that require a supervision signal instructing them how to use their memory cells, the introduction of the \emph{neural Turing machine} \cite{graves2014} allows end-to-end training without this supervision signal, via the use of a content-based soft attention mechanism \cite{goodfellow2016}. Then, \emph{end-to-end memory network} \cite{sukhbaatar2015} that is a form of \emph{memory network} based on a recurrent attention mechanism is proposed.

In these years, an attention mechanism called  \emph{self-attention}, sometimes called \emph{intra-attention}, was successfully implemented within a neural network architecture namely \emph{Long Short-Term Memory-Networks (LSTMN)} \cite{cheng2016}. It modifies the standard LSTM structure by replacing the memory cell with a memory network \cite{weston2014}. This is because memory networks have a set of key vectors and a set of value vectors, whereas LSTMs maintain a hidden vector and a memory vector \cite{cheng2016}. In contrast to attention idea in \cite{bahdanau2015}, memory and attention are added \emph{within} a sequence encoder in LSTMN. In order to compute a representation of a sequence, self-attention is described as relating different positions of it \cite{vaswani2017}. One of the first approaches of self-attention is applied for natural language inference \cite{parikh2016}. 

Many attention-based models have been proposed for neural image captioning \cite{you2016}, abstractive sentence summarization \cite{rush2015}, speech recognition \cite{yu2016,chorowski2015}, automatic video captioning \cite{zanfir2016}, neural machine translation \cite{cheng2016a}, and recognizing textual entailment \cite{rocktaschel2016}. Different attention-based models perform visual question answering \cite{zhu2016,chen2015,xu2016}. An attention-based CNN is presented for modeling sentence pairs \cite{yin2016}. A recurrent soft attention based model learns to focus selectively on parts of the video frames and classifies videos \cite{sharma2016}.  

On the other side, several neural network architectures have been presented in a variety of tasks. For instance, \emph{Stacked Attention Network (SAN)} is described for image question answering \cite{yang2016}. {\em Deep Attention Recurrent Q-Network (DARQN)} integrates soft and hard attention mechanisms into the structure of Deep Q-Network (DQN) \cite{sorokin2015}. {\em Wake-Sleep Recurrent Attention Model (WS-RAM)} speeds up the training time for image classification and caption generation tasks \cite{ba2015}. {\em alignDRAW} model, an extension of the \emph{Deep Recurrent Attention Writer (DRAW)} \cite{gregor2015}, is a generative model of images from captions using a soft attention mechanism \cite{mansimov2016}. \emph{Generative Adversarial What-Where Network (GAWWN)} synthesizes images given instructions describing what content to draw in which location \cite{reed2016}. 

\section{The Transformer: Return of the Attention}
\label{sec:4}
After the proposed attention mechanisms in 2015, researchers published studies that mostly modifying or implementing them to different tasks. However, in 2017, a novel neural network architecture, namely the \emph{Transformer}, based entirely on \emph{self-attention} was presented \cite{vaswani2017}. The Transformer achieved great results on two machine translation tasks in addition to English constituency parsing. The most impressive point about this architecture is that it contains neither recurrence nor convolution. The Transformer performs well by replacing the conventional recurrent layers in encoder-decoder architecture used for NMT with self-attention.     

The Transformer is composed of encoder-decoder stacks each of which has six identical layers within itself. In Fig. \ref{transformer}, one encoder-decoder stack is shown to illustrate the model \cite{vaswani2017}. Each stack includes only attention mechanisms and feedforward neural networks. As this architecture does not include any recurrent or convolutional layer, information about the relative or absolute positions in the input sequence is given at the beginning of both encoder and decoder using {\emph{positional encodings}.

\begin{figure}[h!]
\centering
\includegraphics[scale=.24]{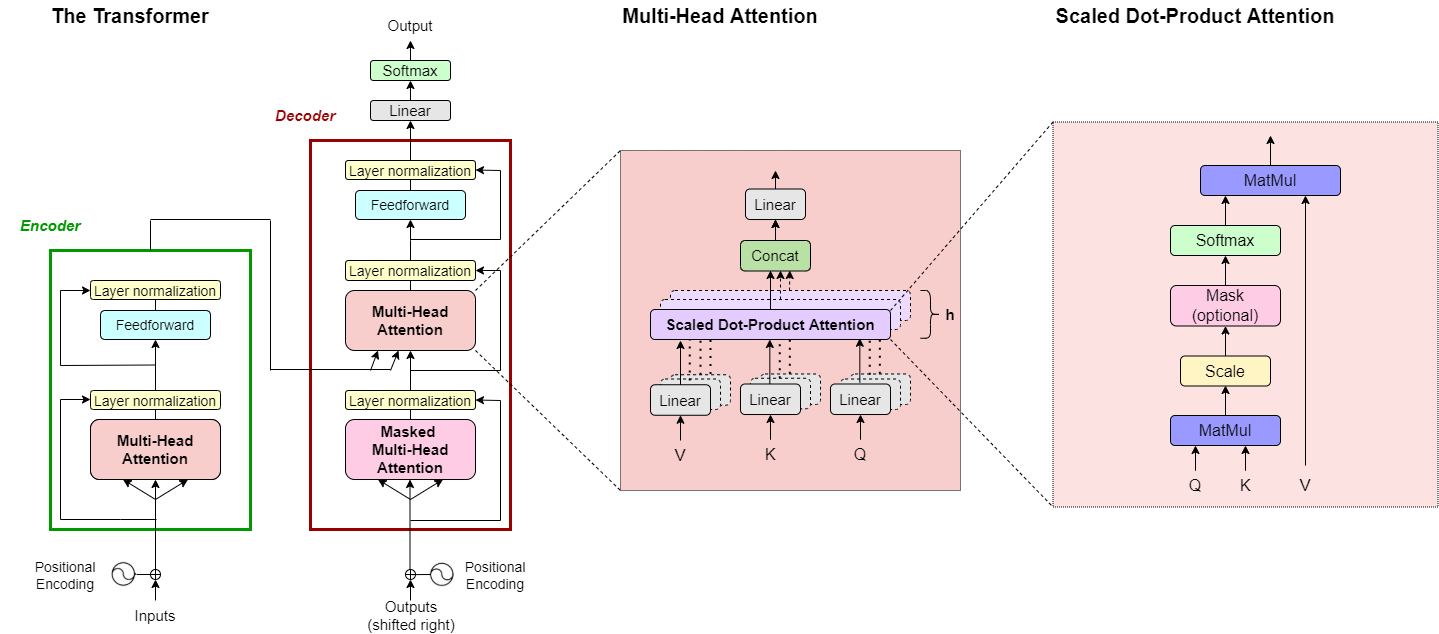}
\caption{The Transformer architecture and the attention mechanisms it uses in detail \cite{vaswani2017}. \emph{(Left)} The Transformer with one encoder-decoder stack. \emph{(Center)} Multi-head attention. \emph{(Right)} Scaled dot-product attention.}
\label{transformer}
\end{figure}

The calculations of self-attention are slightly different from the mechanisms described so far in this paper. It uses three vectors namely {\emph{query}}, {\emph{key}} and {\emph{value}} for each word. These vectors are computed by multiplying the input with weight matrices {\emph{$W_q$}}, {\emph{$W_k$}} and {\emph{$W_v$}} which are learned during training. In general, each value is weighted by a function of the query with the corresponding key. The output is computed as a weighted sum of the values. Based on this idea, two attention mechanisms are proposed: In the first one, called \emph{scaled dot-product attention}, the dot products of the query with all keys are computed as given in the right side of Fig. \ref{transformer}. Each result is divided to the square root of the dimension of the keys to have more stable gradients. They pass into the softmax function, thus the weights for the values are obtained. Finally each softmax score is multiplied with the value as given in Eq. (\ref{selfatt}). The authors propose computing the attention on a set of queries simultaneously by taking queries and keys of dimension $d_k$, and values of dimension $d_v$ as inputs. The keys, queries and values are packed together into matrices \emph{K}, \emph{Q} and \emph{V}. Finally, the output matrix is obtained as follows \cite{vaswani2017}:

\begin{equation}
Attention(Q,K,V) = softmax(\frac{QK^T}{\sqrt{d_k}})V
\label{selfatt}
\end{equation} 

This calculation is performed by every word against the other words. This leads to having {\emph{values}} of each word relative to each other. For instance, if the word $x_2$ is not relevant for the word $x_1$, then the softmax score gives low probability scores. As a result, the corresponding value is decreased. This leads to an increase in the value of relevant words, and those of others decrease. In the end, every word obtains a new value for itself.

As seen from Fig. \ref{transformer}, the Transformer model does not directly use scaled dot-product attention. But the attention mechanism it uses is based on these calculations. The second mechanism proposed, called the \emph{multi-head attention}, linearly projects the queries, keys and values {\em h} times with different, learned linear projections to $d_q$, $d_k$ and $d_v$ dimensions, respectively \cite{vaswani2017}. The attention function is performed in parallel on each of these projected versions of queries, keys and values, i.e., {\emph{heads}}. By this way, $d_v$-dimensional output values are obtained. In order to get the final values, they are concatenated and projected one last time as shown in the center of Fig. \ref{transformer}. By this way, the self-attention is calculated multiple times using different sets of query, key and value vectors. Thus, the model can jointly attend to information at different positions \cite{vaswani2017}:  

\begin{eqnarray}
MultiHead(Q,K,V) = Concat(head_1,...,head_h)W^O  \\
\nonumber    where \; head_i = Attention(QW_i^Q, KW_i^K, VW_i^V)
\end{eqnarray}

In the decoder part of the Transformer, {\emph{masked multi-head attention} is applied first to ensure that only previous word embeddings are used when trying to predict the next word in the sentence. Therefore, the embeddings that shouldn't be seen by the decoder are masked by multiplying with zero.

An interesting study examines the contribution made by individual attention heads in the encoder \cite{voita2019}. Also, there is an evaluation of the effects of self-attention on gradient propagation in recurrent networks \cite{kerg2020}. For a deeper analysis of multi-head self-attention mechanism from a theoretical perspective see \cite{cordonnier2020}.

Self-attention has been used successfully in a variety of tasks including sentence embedding \cite{lin2017} and abstractive summarization \cite{paulus2018}. It is shown that self-attention can lead to improvements to discriminative constituency parser \cite{kitaev2018}, and speech recognition as well \cite{povey2018,vyas2020}. Also, the \emph{listen-attend-spell} model \cite{chan2016} has been improved with the self-attention for acoustic modeling \cite{sperber2018}. 

As soon as these self-attention mechanisms were proposed, they have been incorporated with deep neural networks for a wide range of tasks. For instance, a deep learning model learned a number of large-scale tasks from multiple domains with the aid of self-attention mechanism \cite{kaiser2017}. Novel self-attention neural models are proposed for cross-target stance classification \cite{xu2018a} and NMT \cite{maruf2019}. Another study points out that a fully self-attentional model can reach competitive predictive performance on ImageNet classification and COCO object detection tasks \cite{ramachandran2019}. Besides, developing novel attention mechanisms has been carried out such as \emph{area attention}, a novel mechanism that can be used along multi-head attention \cite{li2019}. It attends to areas in the memory by defining the key of an area as the mean vector of the key of each item, and defining the value as the sum of all value vectors in the area.   

When a novel mechanism is proposed, it is inevitable to incorporate it into the GAN framework \cite{goodfellow2014}. \emph{Self-Attention Generative Adversarial Networks (SAGANs)} \cite{zhang2019} introduce a self-attention mechanism into convolutional GANs. Different from the traditional convolutional GANs, SAGAN generates high-resolution details using cues from all feature locations. Similarly, \emph{Attentional Generative Adversarial Network (AttnGAN)} is presented for text to image generation \cite{xu2018}. On the other side, a machine reading and question answering architecture called {\em QANet} \cite{yu2018} is proposed without any recurrent networks. It uses self-attention to learn the global interaction between each pair of words whereas convolution captures the local structure of the text. In another study, \emph{Gated Attention Network (GaAN)} controls the importance of each attention head's output by introducing gates \cite{zhang2018a}. Another interesting study introduces \emph{attentive group convolutions} with a generalization of visual self-attention \cite{romero2020}. A deep transformer model is implemented for language modeling over long sequences \cite{alrfou2019}.  

\begin{table*}
{\setlength{\tabcolsep}{3pt}
\caption{Summary of Notation}
\begin{center}
\vspace{-2mm}
\begin{tabular}{rccc}
\hline\hline
Symbol & Definition \\
\hline
a   & annotation   \\
c   & context vector  \\
$\alpha$  & weight  \\
e  & energy  \\
f & feedforward neural network  \\
h & hidden state  \\
$\phi$ & hard (stochastic) / soft (deterministic) attention  \\
s & location variable  \\
p & source position \\
$K$, $Q$, $V$ & keys, queries and values matrices, respectively \\
$W_q$, $W_k$, $W_v$ & weight matrices for queries, keys and values, respectively \\  
\hline
\end{tabular}
\vspace{-2mm}
\end{center}
\label{table}}
\end{table*}

\subsection{Self-attention variants}

In recent years, self-attention has become an important research direction within the deep learning community. Self-attention idea has been examined in different aspects. For example, self-attention is handled in a multi-instance learning framework \cite{du2018}. The idea of \emph{Sparse Adaptive Connection (SAC)} is presented for accelerating and structuring self-attention \cite{li2020a}. The research on improving self-attention continues as well \cite{yang2019,yang2018a,bello2019}. Besides, based on the self-attention mechanisms proposed in the Transformer, important studies that modify the self-attention have been presented. Some of the most recent and prominent studies are summarized below.

\paragraph{\textbf{Relation-aware self-attention}}

It extends the self-attention mechanism by regarding representations of the relative positions, or distances between sequence elements \cite{shaw2018}. Thus, it can consider the pairwise relationships between input elements. This type of attention mechanism defines vectors to represent the edge between two inputs. It provides learning two distinct edge representations that can be shared across attention heads without requiring additional linear transformations. 

\paragraph{\textbf{Directional self-attention (DiSA)}} A novel neural network architecture for learning sentence embedding named \emph{Directional Self-Attention Network (DiSAN)} \cite{shen2017} uses \emph{directional} self-attention followed by a \emph{multi-dimensional} attention mechanism. Instead of computing a single importance score for each word based on the word embedding,  multi-dimensional attention computes a feature-wise score vector for each token. To extend this mechanism to the self-attention, two variants are presented: The first one, called \emph{multi-dimensional `token2token' self-attention} generates context-aware coding for each element. The second one, called \emph{multi-dimensional `source2token' self-attention} compresses the sequence into a vector \cite{shen2017}. On the other side, directional self-attention produces context-aware representations with temporal information encoded by using positional masks. By this way, directional information is encoded. First, the input sequence is transformed to a sequence of hidden states by a fully connected layer. Then, multi-dimensional token2token self-attention is applied to these hidden states. Hence, context-aware vector representations are generated for all elements from the input sequence. 

\paragraph{\textbf{Reinforced self-attention (ReSA)}}

A sentence-encoding model named \emph{Reinforced Self-Attention Network (ReSAN)} uses \emph{reinforced self-attention (ReSA)} that integrates soft and hard attention mechanisms into a single model. ReSA selects a subset of head tokens, and relates each head token to a small subset of dependent tokens to generate their context-aware representations \cite{shen2018}. For this purpose, a novel hard attention mechanism called \emph{reinforced sequence sampling (RSS)}, which selects tokens from an input sequence in parallel and trained via policy gradient, is proposed. Given an input sequence, RSS generates an equal-length sequence of binary random variables that indicates both the selected and discarded ones. On the other side, the soft attention provides reward signals back for training the hard attention. The proposed RSS provides a sparse mask to self-attention. ReSA uses two RSS modules to extract the sparse dependencies between each pair of selected tokens. 

\paragraph{\textbf{Outer product attention (OPA)}}

\emph{Self-Attentive Associative Memory (SAM)} is a novel operator based upon \emph{outer product attention (OPA)} \cite{le2020}. This attention mechanism is an extension of dot-product attention \cite{vaswani2017}. OPA differs using element-wise multiplication, outer product, and \emph{tanh} function instead of \emph{softmax}.  

\paragraph{\textbf{Bidirectional block self-attention (Bi-BloSA)}}

Another mechanism, \emph{bidirectional block self-attention (Bi-BloSA)} which is simply a \emph{masked block self-attention (mBloSA)} with forward and backward masks to encode the temporal order information is presented \cite{shen2018a}. Here, mBloSA is composed of three parts from its bottom to top namely \emph{intra-block self-attention}, \emph{inter-block self-attention} and \emph{the context fusion}. It splits a sequence into several length-equal blocks, and applies an intra-block self-attention to each block independently. Then, inter-block self-attention processes the outputs for all blocks. This stacked self-attention model results a reduction in the amount of memory compared to a single one applied to the whole sequence. Finally, a feature fusion gate combines the outputs of intra-block and inter-block self-attention with the original input, to produce the final context-aware representations of all tokens.    

\paragraph{\textbf{Fixed multi-head attention}} 

The \emph{fixed multi-head attention} proposes fixing the head size of the Transformer in the aim of improving the representation power \cite{bhojanapalli2020}. This study emphasizes its importance by setting the head size of attention units to input sequence length.

\paragraph{\textbf{Sparse sinkhorn attention}}

It is based on the idea of differentiable sorting of internal representations \emph{within} the self-attention module \cite{tay2020}. Instead of allowing tokens to only attend to tokens within the same block, it operates on block sorted sequences. Each token attends to tokens in the {\em sorted} block. Thus, tokens that may be far apart in the unsorted sequence can be considered. Additionally, a variant of this mechanism named \emph{SortCut sinkhorn attention} applies a post-sorting truncation of the input sequence.

\paragraph{\textbf{Adaptive attention span}}

\emph{Adaptive attention span} is proposed as an alternative to self-attention \cite{sukhbaatar2019}. It learns the attention span of each head independently. To this end, a masking function inspired by \cite{jernite2017} is used to control the attention span for each head. The purpose of this novel mechanism is to reduce the computational burden of the Transformer.  
Additionally, \emph{dynamic attention span} approach is presented to dynamically change the attention span based on the current input as an extension \cite{luong2015,shu2017}.  

\subsection{Transformer variants}

Different from developing novel self-attention mechanisms, several studies have been published in the aim of improving the performance of the Transformer. These studies mostly modify the model architecture. For instance, an additional recurrence encoder is preferred to model recurrence for Transformer directly \cite{hao2019}. In another study, a new weight initialization scheme is applied to improve Transformer optimization \cite{huang2020}. A novel positional encoding scheme is used to extend the Transformer to tree-structured data \cite{shiv2019}. Investigating model size by handling Transformer width and depth for efficient training is also an active research area \cite{li2020}. Transformer is used in reinforcement learning settings \cite{hoshen2017,hu2021,parisotto2021} and for time series forecasting in adversarial training setting \cite{wu2020a}.  

Besides, many Transformer variants have been presented in the recent past. \emph{COMmonsEnse Transformer (COMET)} is introduced for automatic construction of commonsense knowledge bases \cite{bosselut2019}. \emph{Evolved Transformer} applies neural architecture search for a better Transformer model \cite{so2019}. \emph{Transformer Autoencoder} is a sequential autoencoder for conditional music generation \cite{choi2020}. \emph{CrossTransformer} takes a small number of labeled images and an unlabeled query, and computes distances between spatially-corresponding features to infer class membership \cite{doersch2020}.  \emph{DEtection TRansformer (DETR)} is a new design for object detection systems \cite{carion2020}, and \emph{Deformable DETR} is an improved version that achieves better performance in less time \cite{zhu2021}. \emph{FLOw-bAsed TransformER (FLOATER)} emphasizes the importance of position encoding in the Transformer, and models the position information via a continuous dynamical model \cite{liu2020}.  \emph{Disentangled Context (DisCo) Transformer} simultaneously generates all tokens given different contexts by predicting every word in a sentence conditioned on an arbitrary subset of the rest of the words \cite{kasai2020}. \emph{Generative Adversarial Transformer (GANsformer)} is presented for visual generative modeling \cite{hudson2021}.

Recent work has demonstrated significant performance on NLP tasks. In \emph{OpenAI GPT}, there is a left-to-right architecture, where every token can only attend to previous tokens in the self-attention layers of the Transformer \cite{radford2018}. \emph{GPT-2} \cite{radford2019} and \emph{GPT-3} \cite{brown2020} models have improved the progress. In addition to these variants, some prominent Transformer-based models are summarized below.     

\paragraph{\textbf{Universal Transformer}}

A generalization of the Transformer model named the \emph{Universal Transformer} \cite{dehghani2018} iteratively computes representations $H^t$ at step {\em t} for all positions in the sequence in parallel. To this end, it uses the scaled dot-product attention in Eq. (\ref{selfatt}) where {\em d} is the number of columns of Q, K and V. In the Universal Transformer, the multi-head self-attention with {\em k} heads is used. The representations $H^t$ is mapped to queries, keys and values with affine projections using learned parameter matrices $W^Q \in \Re^{d\times d/k}$, $W^K \in \Re^{d\times d/k}$, $W^V \in \Re^{d\times d/k}$ and $W^O \in \Re^{d\times d}$ \cite{dehghani2018}:    

\begin{eqnarray}
MultiHead(H^t) = Concat(head_1,...,head_k)W^O  \\
\nonumber where \; head_i = Attention(H^tW_i^Q, H^tW_i^K, H^tW_i^V)
\label{multiheadselfuniversaltransformer}    
\end{eqnarray}

\paragraph{\textbf{Image Transformer}}

\emph{Image Transformer} \cite{parmar2018} demonstrates that self-attention based models can also be well-suited for images instead of text. This Transformer type restricts the self-attention mechanism to attend to local neighborhoods. Thus, the size of images that the model can process is increased. Its larger receptive fields allow the Image Transformer to significantly improve the model performance on image generation as well as image super-resolution.

\paragraph{\textbf{Transformer-XL}}

This study aims to improve the fixed-length context of the Transformer \cite{vaswani2017} for language modeling. \emph{Transformer-XL} \cite{dai2019} makes modeling very long-term dependency possible by reusing the hidden states obtained in previous segments. Hence, information can be propagated through the recurrent connections. In order to reuse the hidden states without causing temporal confusion, Transformer-XL uses relative positional encodings. Based on this architecture, a modified version named \emph{the Gated Transformer-XL (GTrXL)} is presented in the reinforcement learning setting \cite{parisotto2020}.

\paragraph{\textbf{Tensorized Transformer}}

\emph{Tensorized Transformer} \cite{ma2019} compresses the multi-head attention in Transformer. To this end, it uses a novel self-attention model \emph{multi-linear attention} with Block-Term Tensor Decomposition (BTD) \cite{lathauwer2008}. It builds a \emph{single-block attention} based on the Tucker decomposition \cite{tucker1966}. Then, it uses a multi-linear attention constructed by a BTD to compress the multi-head attention mechanism. In Tensorized Transformer, the factor matrices are shared across multiple blocks.

\paragraph{\textbf{BERT}}

The \textbf{B}idirectional \textbf{E}ncoder \textbf{R}epresentations from \textbf{T}ransformers \emph{(BERT)} 
aims to pre-train deep bidirectional representations from unlabeled text \cite{devlin2019}. BERT uses a multilayer bidirectional Transformer as the encoder. Besides, inspired by the Cloze task \cite{taylor1953}, it has a \emph{masked language model} pre-training objective. BERT randomly masks some of the tokens from the input, and predicts the original vocabulary id of the masked word based only on its context. This model can pre-train a deep bidirectional Transformer. In all layers, the pre-training is carried out by jointly conditioning on both left and right context. BERT differs from the left-to-right language model pre-training from this aspect. 

Recently, BERT model has been examined in detail. For instance, the behaviour of attention heads are analysed \cite{clark2019}. Various methods have been investigated for compressing \cite{sun2019,wang2020a}, pruning \cite{mccarley2020}, and quantization \cite{zafrir2019}. Also, BERT model has been considered for different tasks such as coreference resolution \cite{joshi2019b}. A novel method is proposed in order to accelerate BERT training \cite{gong2019}.     

Furthermore, various BERT variants have been presented. \emph{ALBERT} aims to increase the training speed of BERT, and presents two parameter reduction techniques \cite{lan2020}. Similarly, \emph{PoWER-BERT} \cite{goyal2020} is developed to improve the inference time of BERT. This scheme is also used to accelerate ALBERT. Also, \emph{TinyBERT} is proposed to accelerate inference and reduce model size while maintaining accuracy \cite{jiao2019}. In order to obtain better representations, \emph{SpanBERT} is proposed as a pre-training method \cite{joshi2020}. As a robustly optimized BERT approach, \emph{RoBERTa} shows that BERT was significantly undertrained \cite{liu2019}. Also, \emph{DeBERTa} improves RoBERTa using the disentangled attention mechanism \cite{he2021}. On the other side, \emph{DistilBERT} shows that it is possible to reach similar performances using much smaller language models pre-trained with knowledge distillation \cite{sanh2019}. \emph{StructBERT} proposes two novel linearization strategies \cite{wang2020}. \emph{Q-BERT} is introduced for quantizing BERT models \cite{shen2020}, \emph{BioBERT} is for biomedical text mining \cite{lee2020}, and  
\emph{RareBERT} is for rare disease diagnosis \cite{prakash2021}.
 
Since 2017 when the Transformer was presented, research directions have generally focused on novel self-attention mechanisms, adapting the Transformer for various tasks, or making them more understandable. In one of the most recent studies, NLP becomes possible in the mobile setting with {\emph{Lite Transformer}. It applies {\emph{long-short range attention} where some heads specialize in the local context modeling while the others specialize in the long-distance relationship modeling \cite{wu2020}. A deep and light-weight Transformer \emph{DeLighT} \cite{mehta2021} and a hypernetwork-based model namely \emph{HyperGrid Transformers} \cite{tay2021} perform with fewer parameters. \emph{Graph Transformer Network} is introduced for learning node representations on heterogeneous graphs    \cite{yun2019} and different applications are performed for molecular data \cite{rong2020} or textual graph representation \cite{yang2021}. Also, {\emph{Transformer-XH} applies {\emph{eXtra Hop attention} for structured text data \cite{zhao2020}. \emph{AttentionXML} is a tree-based model for extreme multi-label text classification \cite{you2019}. Besides, attention mechanism is handled in a Bayesian framework \cite{fan2020}. For a better understanding of Transformers, an identifiability analysis of self-attention weights is conducted in addition to presenting {\emph{effective attention} to improve explanatory interpretations \cite{brunner2020}. Lastly, \emph{Vision Transformer (ViT)} processes an image using a standard Transformer encoder as used in NLP by interpreting it as a sequence of patches, and performs well on image classification tasks \cite{dosovitskiy2021}.

\subsection{What about complexity?}

All these aforementioned studies undoubtedly demonstrate significant success. But success not make one great. The Transformer also brings a very high computational complexity and memory cost. The necessity of storing attention matrix to compute the gradients with respect to queries, keys and values causes a non-negligible quadratic computation and memory requirements. Training the Transformer is a slow process for very long sequences because of its quadratic complexity. There is also time complexity which is quadratic with respect to the sequence length. In order to improve the Transformer in this respect, recent studies have been conducted to improve this issue. One of them is \emph{Linear Transformer} which expresses the self-attention as a linear dot-product of kernel feature maps \cite{katharopoulos2020}. Linear Transformer reduces both memory and time complexity by changing the self-attention from the softmax function in Eq. (\ref{selfatt}) to a feature map based dot-product attention. Its performance is competitive with the vanilla Transformer architecture on image generation and automatic speech recognition tasks while being faster during inference. On the other side, \emph{FMMformers} which use the idea of the \emph{fast multipole method (FMM)} \cite{leslie1987} outperform the linear Transformer by decomposing the attention matrix into near-field and far-field attention with linear time and memory complexity \cite{nguyen2021}.

Another suggestion made in response to the Transformer's quadratic nature is The \emph{Reformer} that replaces dot-product attention by one that uses \emph{locality-sensitive hashing} \cite{kitaev2020}. It reduces the complexity but one limitation of the Reformer is its requirement for the queries and keys to be identical. \emph{Set Transformer} aims to reduce computation time of self-attention from quadratic to linear by using an attention mechanism based on sparse Gaussian process literature \cite{lee2019}. \emph{Routing Transformer} aims to reduce the overall complexity of attention by learning dynamic sparse attention patterns by using \emph{routing attention with clustering} \cite{roy2020}. It applies k-means clustering to model sparse attention matrices. At first, queries and keys are assigned to clusters. The attention scheme is determined by considering only queries and keys from the same cluster. Thus, queries are routed to keys belonging to the same cluster \cite{roy2020}.

\emph{Sparse Transformer} introduces sparse factorizations of the attention matrix by using \emph{factorized self-attention}, and avoids the quadratic growth of computational burden \cite{child2019}. It also shows the possibility of modeling sequences of length one million or more by using self-attention in theory. In the Transformer, all the attention heads with the softmax attention assign a non-zero weight to all context words. \emph{Adaptively Sparse Transformer} replaces softmax with $\alpha$-entmax which is a differentiable generalization of softmax allowing low-scoring words to receive precisely zero weight \cite{correia2019}. By means of context-dependent sparsity patterns, the attention heads become flexible in the Adaptively Sparse Transformer. \emph{Random feature attention} approximates softmax attention with random feature methods \cite{peng2021}. \emph{Skyformer} replaces softmax with a Gaussian kernel and adapts Nyström method \cite{chen2021}.  A sparse attention mechanism named \emph{BIGBIRD} aims to reduce the quadratic dependency of Transformer-based models to linear \cite{zaheer2020}. Different from the similar studies, BIGBIRD performs well for genomics data alongside NLP tasks such as question answering. 

\emph{Music Transformer} \cite{huang2019} shows that self-attention can also be useful for modeling music. This study emphasizes the infeasibility of the relative position representations introduced by \cite{shaw2018} for long sequences because of the quadratic intermediate relative information in the sequence length. Therefore, this study presents an extended version of relative attention named \emph{relative local attention} that improves the relative attention for longer musical compositions by reducing its intermediate memory requirement to linear in the sequence length. A softmax-free Transformer (\emph{SOFT}) is presented to improve the computational efficiency of ViT. It uses Gaussian kernel function instead of the dot-product similarity \cite{lu2021}.

Additionally, various approaches have been presented in {\emph{Hierarchical Visual Transformer} \cite{pan2021}, \emph{Long-Short Transformer (Transformer-LS)} \cite{zhu2021a}, \emph{Perceiver} \cite{jaegle2021}, and  {\emph{Performer}} \cite{choromanski2021}. Image Transformer based on the cross-covariance matrix between keys and queries is applied \cite{elnouby2021}, and a new vision Transformer is proposed \cite{yu2021}. Furthermore, a Bernoulli sampling attention mechanism decreases the quadratic complexity to linear \cite{zeng2021}. A novel linearized attention mechanism performs well on object detection, instance segmentation, and stereo depth estimation \cite{shen2021}. A study shows that kernelized attention with relative positional encoding can be calculated using Fast Fourier Transform and it leads to get rid of the quadratic complexity for long sequences \cite{luo2021}. A linear unified nested attention mechanism namely \emph{Luna} uses two nested attention functions to approximate the softmax attention in Transformer to achieve linear time and space complexity \cite{ma2021}.

\section{Concluding Remarks: A New Hope} 
\label{conc}
Inspired by the human visual system, the attention mechanisms in neural networks have been developing for a long time. In this study, we examine this duration beginning with its roots up to the present time. Some mechanisms have been modified, or novel mechanisms have emerged in this period. Today, this journey has reached a very important stage. The idea of incorporating attention mechanisms into deep neural networks has led to state-of-the-art results for a large variety of tasks. Self-attention mechanisms and \emph{GPT-n} family models have become a new hope for more advanced models. These promising progress bring the questions whether the attention could help further development, replace the popular neural network layers, or could be a better idea than the existing attention mechanisms? It is still an active research area and much to learn we still have, but it is obvious that more powerful systems are awaiting when neural networks and attention mechanisms join forces.

%
%


%
 \section*{Conflict of interest}
The author declares that she has no conflict of interest.

\bibliographystyle{spphys}       
\bibliography{mybibfile}   

%
%


\end{document}